\documentclass[review]{elsarticle}
\usepackage{times}
\usepackage{multirow}
\usepackage{times}
\usepackage{helvet}
\usepackage{courier}
\usepackage{graphicx}
\usepackage{CJK}

\journal{arXiv}

\begin{document}
\begin{CJK*}{UTF8}{gbsn}

\begin{frontmatter}

\title{Neural Personalized Response Generation as \\Domain Adaptation}

\author{Wei-Nan Zhang}
\ead{wnzhang@ir.hit.edu.cn}
\author{Ting Liu}
\author{Yifa Wang}
\author{Qingfu Zhu}
\address{Research Center for Social Computing and Information Retrieval, School of Computer Science and Technology, Harbin Institute of Technology, Harbin, China, 150001}


\begin{abstract}
In this paper, we focus on the personalized response generation for conversational systems.
Based on the sequence to sequence learning, especially the encoder-decoder framework, we propose a two-phase approach, namely initialization then adaptation, to model the responding style of human and then generate personalized responses.
For evaluation, we propose a novel human aided method to evaluate the performance of the personalized response generation models by online real-time conversation and offline human judgement.
Moreover, the lexical divergence of the responses generated by the 5 personalized models indicates that the proposed two-phase approach achieves good results on modeling the responding style of human and generating personalized responses for the conversational systems.
\end{abstract}

\begin{keyword}
\texttt {Personalized Response Generation \sep Conversational Systems \sep Sequence to Sequence Learning \sep Domain Adaptation}
\end{keyword}

\end{frontmatter}


\section{Introduction}

Conversational system, which is also called conversational robot, virtual agent or chatbot, etc, is an interesting and challenging research of artificial intelligence.
It can be applied to a large number of scenarios of human-computer interaction, such as question answering~\cite{1}, negotiation~\cite{2,3}, e-commence~\cite{4}, tutoring~\cite{5}, etc.
Recently, conversational system usually plays the role of virtual companion or assistant of human.
For example, the virtual assistant on mobile phone is one of the most popular application of d system, such as, Siri, Cortana, Facebook M, Viv, etc.

Despite the functional success of the existing conversational systems, literature that tries to model the personalized responding of a conversational system is still sparse.
Table~\ref{tab-intro} shows an example of the responses of different personality to a given post.

\begin{table}
\begin{center}
\caption{\label{tab-intro}An example of the responses of different personality to a given post.}{
\begin{tabular}{l|l}
\hline
Post & Is it a proper dress for the first date? \\
\hline
Response \#1 & Yep. \\
\hline
Response \#2 & Honey, it is very suitable! \\
\hline
Response \#3 & It is better to wearing a silk scarf. \\
\hline
\end{tabular}}
\end{center}
\end{table}

From Table~\ref{tab-intro}, we can see that the Response \#1 is a briefly definite response to the post.
The Response \#2 is an emotional response.
And the Response \#3 gives another suggestion on dressing.
It is obvious that although the 3 responses are relevant to post, they represent the different personalities and language styles respectively that may further impact the chatting process and the user experience.

To address the problem of generating personalized response for conversational systems, in this paper, we proposed an optimized sequence to sequence (seq2seq) learning framework, which is based on the recurrent neural network, to modeling the input post and generating the personalized response.
Unlike the previous seq2seq approaches for response generation that learn to encode and decode in the same way, the proposed approach fully adopts the advantages of the seq2seq approaches and then learns to adapt to generate personalized response.

The contributions of this paper are two-fold:
\begin{itemize}
  \item We proposed a two-phase approach, namely initialization then adaptation, to learn to generate the personalized response for conversational systems.
  \item We proposed a novel human aided method to evaluate the personality of the generated response.
\end{itemize}

The rest of the paper is organized as follows:
We will detail the proposed two-phase approach to generating personalized response in the next section.
In the following section, the experimental results and analysis will be presented.
The last two sections are the related work and conclusion respectively.

\section{Our Approach}\label{app}

Recently, a number of research works have been proposed to generate sentence or response based on the recurrent neural network (RNN)~\cite{6,7,8,9,10,11,12,13,14,15,16,17,18,19,22,53}.
The target of these work is that given an input sentence $x$, the RNN based model tries to generate an output sentence $y$ that can maximize the conditional probability of $p(y|x)$.
For the response generation of a conversational system, this type of approach usually includes two parts, namely the encoder and decoder.
The encoder is to convert the input sentence into a vector which represents the complete semantics of the input sentence.
The decoder then generates the output sentence character by character (or word by word) according to the information on the encoding phase.

\subsection{Personalized Response Generation}

Inspired by the RNN encoder-decoder framework, which is proposed by~\cite{17} and~\cite{6}, in this paper, we proposed a personalized response generation approach for conversational systems.

\begin{figure*}
\centering
\includegraphics[width=0.85\textwidth]{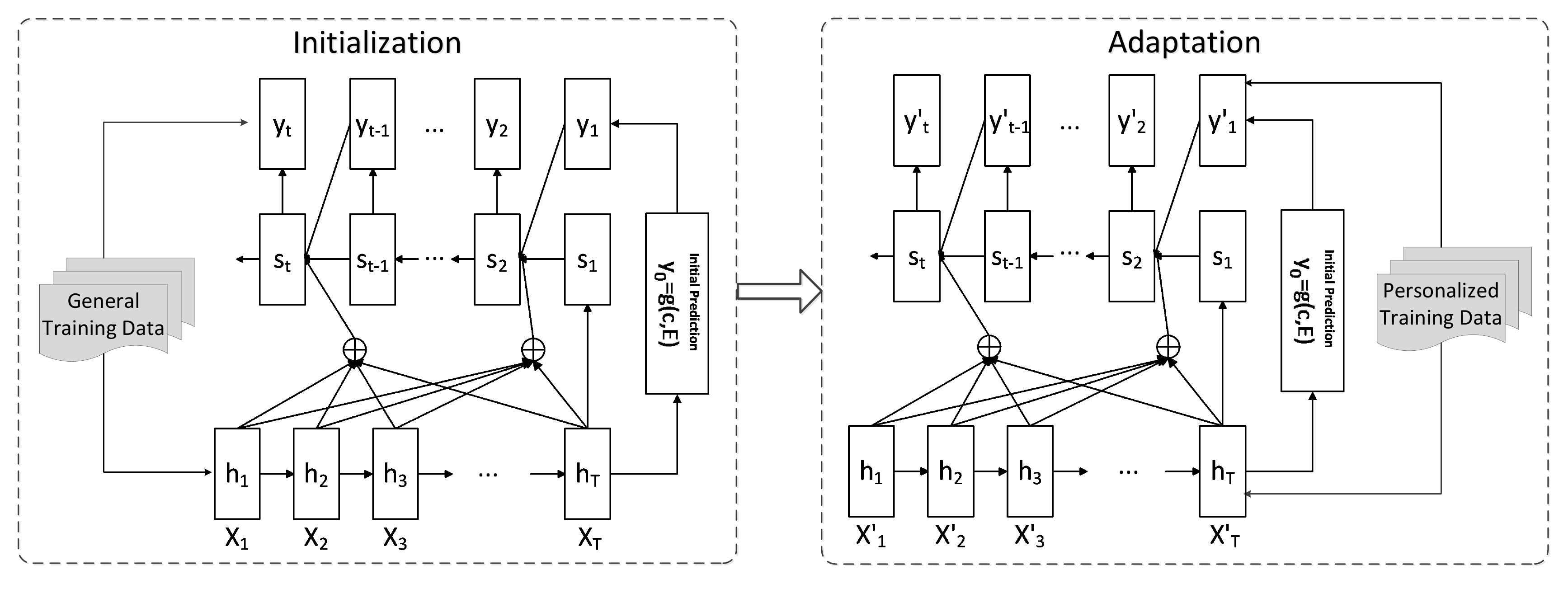}
\caption{\label{fig-app} The framework of the proposed approach.}
\end{figure*}

Figure~\ref{fig-app} shows the framework of the proposed approach.

From Figure~\ref{fig-app}, we can see that the proposed approach consists of two components, namely initialization then adaptation, the first of which pre-trains the responding model of the conversational system using the large scale general training data and the second step fine-tunes the model on the small size of personalized training data.

Next, we will detail the proposed personalized response generation model.
Typically, the encoder and decoder are implemented by the GRU~\cite{20,21} or LSTM~\cite{23,24,25} based RNN.
The encoder reads the input sentence word by word and outputs the hidden state of each word.
These states are denoted as $H$ which is also called annotations.
Here, $h_i$ represents the hidden state at time $i$ and it is computed by its last hidden state $h_{i-1}$ and the input word at time $i$, $X_i$.
Therefore, the hidden state at time $t$ can be denoted as:
\begin{equation}
h_t = f(h_{t-1},X_t);\ \ H=\{h_1,h_2,...,h_T\}
\end{equation}
Here, $T$ equals to the length (the number of words) of the input sentence and $f$ is a non-linear function which can be implemented as LSTM~\cite{6} or GRU~\cite{21}.

The encoder then converts these hidden states to a context vector $c$ as a summary of the semantics of the input sentence.
\begin{equation}
c = q(\{h_1,h_2,...,h_T\})
\end{equation}
Where, $c$ can be implemented in many ways, such as~\cite{6} set $c=h_T$.

For the decoding process, $s_i$ denotes the hidden state at time $i$.
It is also computed by a non-linear function $f$, of which the variables are the output $y_{i-1}$ and the hidden state $s_{i-1}$ at last time. The hidden state of the decoder at time $t$ can be computed as:
\begin{equation}
s_t = f(s_{t-1},y_{t-1})
\end{equation}
Note that the context vector $c$, which is generated from the encoder, is also used to initialize the first hidden state~\cite{6} or all of the hidden states~\cite{17} of the decoder to make sure that the decoder can be conditioned by the encoder.
Therefore, the hidden state of the decoder at time $t$ is updated as:
\begin{equation}
s_t = f(s_{t-1},y_{t-1},c)
\end{equation}
The output of the decoder at the state $s_t$ is to map to a distribution over the vocabulary by using the maxout activation function~\cite{26}

In the proposed approach, the response generation model is implemented by an RNN encoder-decoder framework, which is inspired by~\cite{18}.
For the personalized response generation, we proposed a two-phase training approach to first initialize the model using the large scale general training data and then use the personalized training data to fine-tune the model to generate personalized response.
All the parameters of the responding model are shared between the initialization and adaptation process.

To balance the generation performance and training time of the model, in this paper, we choose the GRU as the non-linear function $f$ for both encoder and decoder.

In this paper, we utilize a weighted sum scheme~\cite{18} to dynamically compute the $c_i$ for each state in the encoding process as:
\begin{equation}\label{att}
c_i=\sum_{j=1}^{T} \alpha_{ij}h_j
\end{equation}
The weight $\alpha_{ij}$ of each hidden state $h_j$ is computed as:
\begin{equation}
\alpha_{ij}=\frac{\exp(e_{ij})}{\sum_{k=1}^{T} \exp(e_{ik})}
\end{equation}
Where, $e_{ij}=a(s_{i-1},h_i)$ is a feedforward neural network, which can be called as the alignment model or attention model~\cite{18}.

\subsection{Generation Quality Optimization}

Compared with the response that is generated by the general RNN encoder-decoder model, the two-phase training approach could generate distinct and personalized response.
However, we found another problem that when the first word is decoded to a high frequency word in the vocabulary, such as ``We'', ``I'', ``Yes'', etc, the personalized response model is tend to generate a general response.

To address the above problem, we utilized a learning scheme to generate the first word in decoding process, namely Learning to Start (LTS) model~\cite{27}.
Unlike the classic RNN encoder-decoder framework that uses a special character ``$<$/s$>$'' to generate the first word in decoding process, the LTS model independently learns to predict the first word using the context vector, which is generated from the encoding process.
The LTS model can be formalized as:
\begin{equation}\label{lts}
y_0=\sigma((\sigma(W_ic)+b_i)E+b_e)
\end{equation}
Here, $c$ is the context vector which is computed by Equation (\ref{att}).
$E$ represents the word embedding matrix of the decoder, $b_i$ and $b_e$ are bias items.
$W_i$ is a learnable matrix that is trained to model the conditional dependence of the context vector $c$ and the first word in decoding process.

By ignoring the bias items, the Equation (\ref{lts}) can be transformed as follows:
\begin{equation}
y_0=g(c,E)
\end{equation}
We thus found that the LTS is to model the relation between the context vector that is generated from the encoder and the embedding matrix of the decoder.
According to the distribution of the generation probability over the decoding vocabulary, LTS predicts the first word for the decoder and then the decoding process goes on until the response is generated completely.
In this paper, we use different vocabularies for encoding and decoding to model the personality of the speaker and responder respectively.

\section{Experiments and Analysis}\label{exp}

\subsection{Experiment Data and Settings}

As the proposed personalized response generation approach includes two phases, there are two separate training data sets, namely general training data and personalized training data (See Figure~\ref{fig-app}).
The data for general training is collected from several Chinese online forums.
It contains 1 million one-to-one post and response pairs and the vocabulary contains 35 thousands words.
Here, one-to-one means one post is only corresponded to one response.
In this paper, LTP (http://www.ltp-cloud.com/) toolkit is used for Chinese word segmentation for all the data.

For the personalized training, we invited 5 volunteers.
Each of them shared 2,000 messages of their chatting history from the use of instant messaging service without any privacy information.
The general messages, such as ``你好(Hello)'', ``早上好(Good Morning)'', ``再见(See you)'', etc, are removed from the personalized training data.

Note that, to train the personalized responding model, these messages from the chatting history can only be used as responses.
Therefore, we need to collect a unique post for each of the messages.
We built a search engine, which is based on the Lucene\footnote{http://lucene.apache.org/} toolkit, to retrieve similar responses from the general training data for the messages.
For each message, the post of the most similar response to the message was chosen as the post of the message.
Therefore, for each volunteer, we thus obtained 2,000 post and message pairs for personalized training.
After training, we have 5 personalized responding models that correspond to the 5 volunteers for the test.

The parameter settings in the response generation model are as follow:

The dimension of the hidden layer of the encoder and decoder is 1,024.
The dimension of the word embedding is 500 which is obtained by using the word2vec toolkit~\cite{49}.
The word2vec is trained with the SogouCS\&CA corpus (2008 version)\footnote{http://www.sogou.com/labs/dl/cs.html}, which is widely used for Chinese text analysis~\cite{51,52}.
The details of the training corpus for training word embedding are shown in Table~\ref{ds4embedding}.

\begin{table}[htbp]
\begin{center}
\caption{Statistics of the training corpus for word embedding generation.\label{ds4embedding}}
{
\begin{tabular}{l|cccc}
\hline
& Data size & \# of tokens & Vocabulary size & \# of threads \\
\hline
SogouCS\&CA & 8.7GB & 1,520,842,220 & 1,354,247 & 10 \\
\hline
\end{tabular}}
\end{center}
\end{table}

The encoder-decoder framework is implemented by using Theano toolkit~\cite{50}.
The batch size is set to 128.
The iteration times are set to 10 and 8 for the general training and personalized training respectively.

\subsection{Test and Evaluation}

It is a non-trivial task to evaluate the performance of a responding model automatically.
As~\cite{7} said: ``Automatic evaluation of response generation is still an open problem.''.
The BLEU socre~\cite{48}, which is widely used in machine translation, is not a suitable evaluation metric for response generation. As the responses to the same post may share less common words, it is impossible to construct a reference set with adequate coverage.
Meanwhile, the Perplexity, which is an evaluation metric for language modeling, is also not reasonable for evaluate the relevance between post and response.

\begin{figure}[htbp]
\centering
\includegraphics[width=0.6\textwidth]{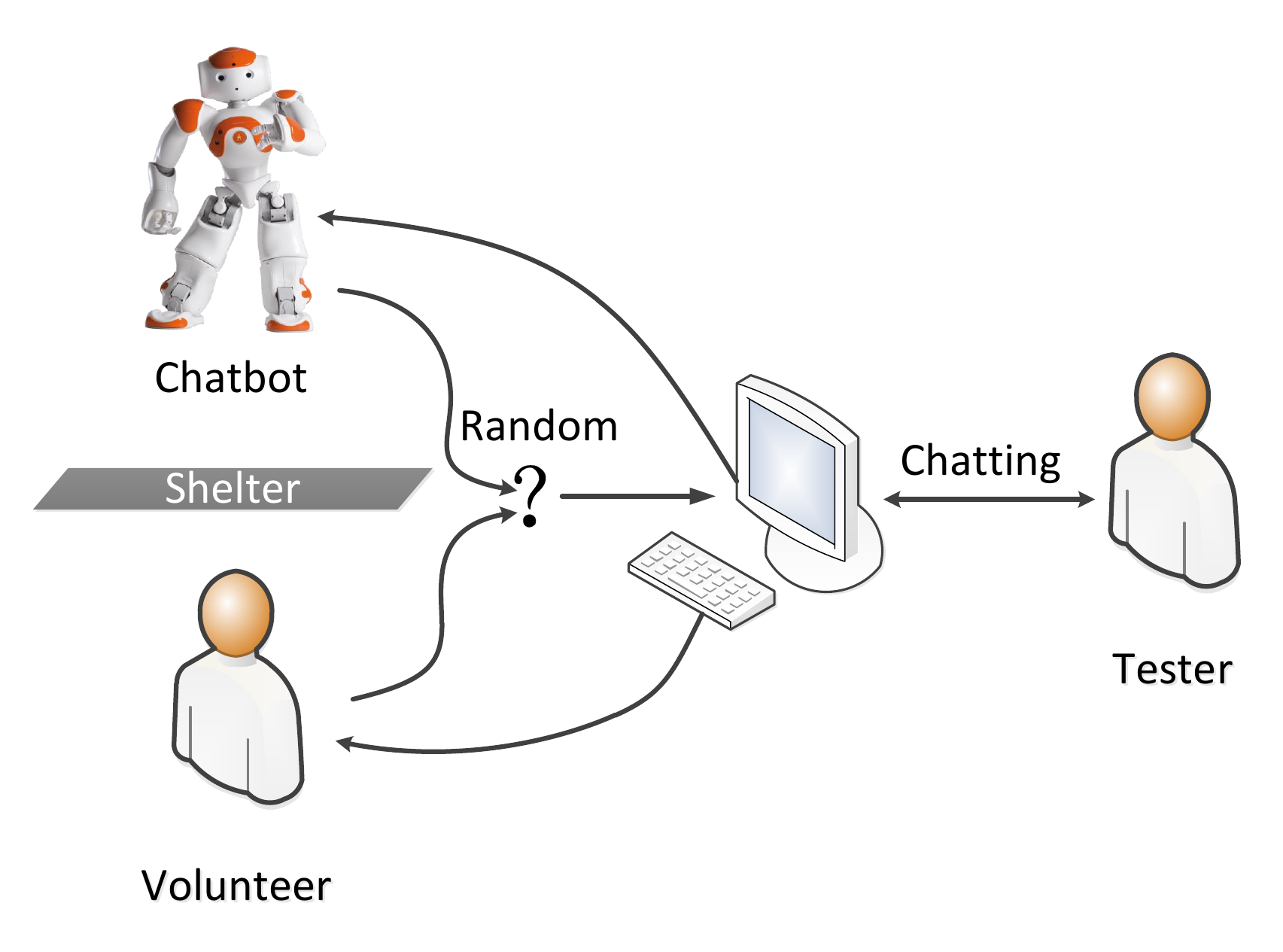}
\caption{\label{hjudge} The evaluation method based on human judgement for personalized response generation.}
\end{figure}

To address the above issues, we design a novel human aided evaluation method based on the online real-time chatting and offline human judgment.
The diagram of the evaluation method is shown in Figure~\ref{hjudge}.

The evaluation method includes a volunteer, a tester and a chatbot.
The volunteer and the tester are communicating through an instant messaging service.
Here, the tester is told to talk to a volunteer through the instant messaging service without any constraints and preconditions as well as the tester do not know the existence of the chatbot.
During the chatting, each of the tester's messages is sent to the volunteer and the chatbot simultaneously.
The \emph{Shelter} in Figure~\ref{hjudge} represents that the volunteer could not see the response that is generated by the chatbot before it is sent to the tester.
The question mark ``\textbf{?}'' denotes that the volunteer needs to randomly decide whether to respond by himself/herself or let the chatbot sends its response.
The reason is to reduce the preference of the volunteer to the response of the chatbot.

When the chatting is finished, the tester is asked to judge whether each of the responses is from the volunteer or someone else.
We defined the \emph{imitation rate}, $r_{imi}$ to evaluate the personality of the proposed dialogue generation approach.
Here, we use $n_{imi}$ to denote the number of responses that are generated by the chatbot, but are judged as the responses of the volunteer.
$n_{gr}$ is the total number of responses that are generated by the chatbot.
The \emph{imitation rate} is thus represented as:
\begin{equation}\label{imi}
r_{imi}=\frac{n_{imi}}{n_{gr}}
\end{equation}

We can obviously see from Equation (\ref{imi}) that the \emph{imitation rate} can reflect the ability of the chatbot on imitating the personalized responding style of the volunteers.
The larger the \emph{imitation rate} is, the better the chatbot imitates the volunteers.

Note that we totally recruit 5 volunteers and 1 tester.
All of them are not participating in the training data collecting and coding.
They also have no experience on developing the relevant techniques.
The tester and the volunteers have known each other very well before the evaluation as well.

\subsection{Experimental Results}

As we described on the above section, we use the \emph{imitation rate} to evaluate the personality of the responses which are generated by the chatbot.
Table~\ref{imi-result} shows the experimental results of the proposed approach to generating personalized responses.

\begin{table*}[htbp]
\begin{center}
\caption{The experimental results of the proposed approach to personalized response generation. $n_{gr}$ and $n_{vr}$ represents the number of responses that are generated by the chatbot and the volunteer respectively. $n_{test}$ is the total number of posts for testing. $n_{imi}$ denotes the number of responses that is generated by the chatbot but are judged as the responses of the volunteer. $r_{imi}$ denotes the imitation rate, which is defined in Equation (\ref{imi}). \label{imi-result}}
{
\begin{tabular}{l|ccccc|c}
\hline
& Volunteer \#1 & Volunteer \#2 & Volunteer \#3 & Volunteer \#4 & Volunteer \#5 & Sum \\
\hline
$n_{gr}$ & 29 & 26 & 21 & 33 & 33 & 142 \\
$n_{vr}$ & 21 & 24 & 29 & 17 & 17 & 106 \\
$n_{test}$ & 50 & 50 & 50 & 50 & 50 & 250 \\
$n_{imi}$ & 11 & 9 & 8 & 13 & 9 & 50 \\
\hline
$r_{imi}$ & 37.93\% & 34.62\% & 38.10\%	& 39.40\% & 27.27\%	& 35.21\% \\
\hline
\end{tabular}}
\end{center}
\end{table*}

Here, the tester sends 50 posts to each volunteer and the 5 test sets shared a small amount of common messages.
Meanwhile, the messages sent to the tester are hybrid responses.
That is to say, in a conversation, the responses to the tester are given by the chatbot and a volunteer in a random inserting way.
Therefore, based on the above conditions, the judgement of the tester on the hybrid responses can be seen as a ``relative'' evaluation on the ability of the chatbot for imitating the personalized responding style of the volunteer.

Moreover, we plan to evaluate the ``absolute'' ability of the chatbot on imitating the personalized responding style of the volunteer. We first randomly sample 50 posts that are sent to the volunteers by the tester in previous evaluation.
We then use the 5 trained models to generate personalized responses of the 50 posts respectively.
Finally, we again ask the tester to judge whether each of the responses is from the volunteer or someone else.
Table~\ref{absolute} shows the results of the human judgement on the 5 personalized response generation models.

\begin{table}[!t]
\begin{center}
\caption{The evaluation results of the 5 personalized response generation models by human judgement. $n_{gr}$ represents the number of responses that are generated by the chatbot. $n_{imi}$ denotes the number of responses that is generated by the chatbot but are judged as the responses of the volunteer. $r_{imi}$ denotes the imitation rate, which is defined in Equation (\ref{imi}). PRM is short for the personalized responding model. \label{absolute}}
{
\begin{tabular}{l|ccccc}
\hline
& PRM \#1 & PRM \#2 & PRM \#3 & PRM \#4 & PRM \#5 \\
\hline
$n_{gr}$ & 50 & 50 & 50 & 50 & 50 \\
$n_{imi}$ & 6 & 8 & 8 & 13 & 10 \\
\hline
$r_{imi}$ & 12\% & 16\% & 16\% & 26\% & 20\% \\
\hline
\end{tabular}}
\end{center}
\end{table}

From Table~\ref{absolute}, we can see that the $r_{imi}$ scores of the 5 personalized responding models (PRM \#1 - \#5) are much lower than those of the corresponding volunteers in Table~\ref{imi-result}.
There are two reasons.
First, on the online real-time chatting, the responses generated by chatbot are randomly inserted to the conversation with the volunteers' responses.
It may increase the confusion of the tester on the judgement.
Second, the online real-time chatting process is context-aware.
Due to the ``coherent model'' in mind, the volunteers may tend to extend the conversation topic.
It may make the tester believe that the previous responses are also typed by the volunteer rather than generated by a chatbot.

\begin{figure}
\centering
\includegraphics[width=0.6\textwidth]{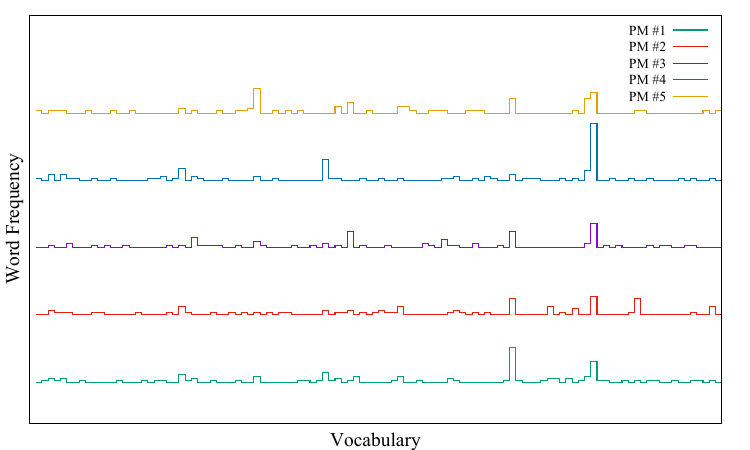}
\caption{\label{distribution} The distributions of the responding words on the vocabulary.}
\end{figure}

To further verify the ability of the PRM \#1 - \#5 on modeling the personalized responding style, we calculate the lexical distributions of the responses generated by the PRM \#1 - \#5 respectively.
Figure~\ref{distribution} shows the distributions of the responding words on the vocabulary.

Here, to calculate the distribution, we combine all the 250 posts of the tester on chatting with the 5 volunteers.
After removing the duplicate posts, we totally obtain 230 posts.
We then utilize the trained models of personalized response generation, PRM \#1 - \#5, to generate the responses of the 230 posts.
The vocabulary size in Figure~\ref{distribution} is 111 after removing the stop words.
From Figure~\ref{distribution}, we can see that the lexical distributions of the responses generated by the 5 personalized models are quite different.
It indicates that through the personalized training, the PRM \#1 - \#5 models can generate the responses with different responding styles, which is embodied in the different lexical distributions.
For an intuitive understanding, Table~\ref{case} shows the real cases of the same post with different responses generated by PRM \#1 - \#5 models.

\begin{table}[htbp]
\begin{center}
\caption{The real cases of the same post with different responses generated by PRM \#1 - \#5 models.\label{case}}
{
\begin{tabular}{l|l|l}
\hline
Post & Models & Responses \\
\hline
\multirow{5}{5cm}{在哪能租到房子？\\(Where can I rent a house?)} & PRM \#1 & 我不知道那个网站。(I don't know the website.) \\
& PRM \#2 & 去找一个吧。(To hunt a house.) \\
& PRM \#3 & 市区。(The downtown.) \\
& PRM \#4 & 联系中介吧。(You can contact the accommodation agency.) \\
& PRM \#5 & 我这有地方住。(I have somewhere to stay.) \\
\hline
\multirow{5}{5cm}{什么编程语言最好？\\(What is the best programming language?)} & PRM \#1 & 日语。(Japanese.) \\
& PRM \#2 & 通常来说是Java。(Generally speaking, it is Java.) \\
& PRM \#3 & 计算机语言。(The computer language.) \\
& PRM \#4 & 不用。(I don't need.) \\
& PRM \#5 & PHP。(PHP.) \\
\hline
\multirow{5}{5cm}{洗衣服好累啊。\\(It is too tired to wash the clothes.)} & PRM \#1 & 用洗衣机洗。(Use the washing machine.) \\
& PRM \#2 & 累。(I'm tired.) \\
& PRM \#3 & 我能。(I can.) \\
& PRM \#4 & 恩。(Yes.) \\
& PRM \#5 & 去休息吧。(Have a rest.) \\
\hline
\end{tabular}}
\end{center}
\end{table}

Next, for each of the personalized responding models, it is also necessary to validate the personality of the generated response towards the corresponding volunteer's.
Therefore, we calculate the percentage of the overlapping words between the volunteer's responses and the chatbot generated responses for a same post set.
We totally obtain 35 common posts from the 5 volunteers.
The percentages of the overlapping words (stop words are removed) are shown in Table~\ref{overlap}.

\begin{table}[!t]
\begin{center}
\caption{The percentage of the overlapping words between the volunteer's responses and the chatbot generated responses for a same post set. ``V'' is short for ``Volunteer''. \label{overlap}}
{
\begin{tabular}{l|ccccc}
\hline
& PRM \#1 & PRM \#2 & PRM \#3 & PRM \#4 & PRM \#5 \\
\hline
V \#1 & \textbf{20.00\%} & 17.86\% & 8.73\% & 14.78\% & 9.65\% \\
V \#2 & 15.45\% & \textbf{18.64\%} & 8.13\% & 12.00\% & 10.83\% \\
V \#3 & 17.65\% & 10.75\% & \textbf{22.22\%} & 14.77\% & 9.57\% \\
V \#4 & 17.63\% & 16.82\% & 8.28\% & \textbf{17.92\%} & 11.32\% \\
V \#5 & 12.12\% & 12.96\% & 6.73\% & 10.28\% & \textbf{18.52\%} \\
\hline
\end{tabular}}
\end{center}
\end{table}

In Table~\ref{overlap}, the ridge of the percentages along the diagonal illustrates the ability of the proposed personalized responding models, PRM \#1 - \#5, on capturing the the personalized responding style of the 5 volunteers.

\section{Related Work}\label{rel}

In this section, we will present the related work of the response generation in two aspects.

\begin{itemize}
  \item \textbf{Task-oriented Dialogue Generation} \\
  The most successful research on the task-oriented dialogue system is mainly based on the partially observed Markov decision process (POMDP)~\cite{28}.
  The task oriented dialogue system mainly focuses on the dialogue state tracking, action classification, policy and reward learning, etc.
  Previous research on task-oriented dialogue generation usually employed handcrafted generator to define the generation decision space with the handcrafted features or statistical models~\cite{34,35,36,37,38,39,40,41}.
  These approaches have the limitation on the scalability to new domains.
  \cite{29} proposed a statistical language generator which used a dynamic Bayesian networks to generate dialogue response.
  \cite{30} learned to generate paraphrases in dialogue through a factored language model that was training from the data collected by crowdsourcing.
  Both of them completely learn from the data and thus has no limitation on domain transfer.
  Recently, as the powerful of deep neural network on learning from large-scale data, \cite{16} proposed a statistical dialogue generator based on a joint recurrent and convolutional neural network, which can directly learn from the data without any semantic alignment or handcrafted rules.
  Further, \cite{15} and \cite{14} proposed a semantically conditioned LSTM to generate dialogue response and then compared it with an RNN encoder-decoder generator on multi-domain data to verify the ability of domain adaptation of the two generators.
  \item \textbf{Non-task-oriented Dialogue Generation} \\
  A non-task-oriented dialogue system is sometimes called an open-domain dialogue system.
  \cite{45} proposed an unsupervised approach to modeling dialogue response by clustering the raw utterances.
  They then presented an end-to-end dialogue response generator by using a phrase-based statistical machine translation model~\cite{42}.
  \cite{44} introduced a search-based system, namely IRIS, to generate dialogues using vector space model and then released the experimental corpus for research and development~\cite{46}.
  \cite{43} introduced the knowledge bases that obtained from the Web to deal with the out-of-domain request.
  Recently, benefit from the advantages of the sequence to sequence learning framework with neural networks~\cite{6}, \cite{7} and \cite{19} had drawn inspiration from the neural machine translation~\cite{17,18} and proposed an RNN encoder-decoder based approach to generate dialogue by considering the last one sentence and a larger range of context respectively.
  \cite{8} presented a hierarchical neural network, which is inspired by~\cite{47}, to build an end-to-end dialogue system.
  \cite{10} and \cite{27} focused on resolving the generating of safe, commonplace, high frequency responses on the sequence to sequence neural network.
  Most recently, \cite{12} captured the advantages of the RNN encoder-decoder on response generation and the deep reinforcement learning on dialogue rewarding to generate context-aware dialogue responses.
\end{itemize}

The most similar work to this paper is~\cite{53}.
They devote to handling the information consistency of a speaker by integrating the speaker embedding and word embedding into a sequence to sequence learning framework.
In this paper, we focus on modeling the personalized responding style of human by presenting a domain adaptation approach, which is also based on the sequence to sequence learning framework.
We believe that the information consistency and responding style are two aspects of the human personality.
We leave the work of jointly exploring the consistency and responding style for personalized dialogue generation in future.

\section{Conclusion}\label{con}

In this paper, we proposed a neural response generation approach, which is based on the encoder-decoder framework, to learn the responding style of human (the volunteers) and then generate personalized response for conversational systems.
The proposed approach extend the traditional encoder-decoder approach to response generation by introducing a domain adaptation scheme, namely initialization-then-adaptation.
We also proposed a novel human aided method to evaluate the ability of the chatbot on imitating the responding styles of the volunteers.
Experimental results on validating the lexical distribution and the word overlapping indicate that the 5 responding models can capture the personalized responding styles of the 5 volunteers.
In future, we will further improve the performance of personalized responding models by introducing the contextual information.
\section*{Acknowledgement}
%
This work is supported by the Fundamental Research Funds for the Central Universities(30620170037).

\section*{References}

\bibliography{elsarticle-template}
\end{CJK*}
\end{document}